\documentclass[sigconf,nonacm]{acmart}

\AtBeginDocument{%
  \providecommand\BibTeX{{%
    \normalfont B\kern-0.5em{\scshape i\kern-0.25em b}\kern-0.8em\TeX}}}

\usepackage{subcaption}

\ExplSyntaxOn
\NewDocumentCommand{\hash}{O{\ttfamily}m}
 {
  \group_begin:
  #1 % font choice, default \ttfamily
  \tl_map_function:nN { #2 } \__anab_hash_char:n
  \group_end:
 }
\cs_new:Nn \__anab_hash_char:n { #1 \skip_horizontal:n { 0pt plus 0.2pt } }
\ExplSyntaxOff

\graphicspath{{./figures/}} 

\begin{document}

%%
%% The "title" command has an optional "short title" to be used in page headers.
\title{Uncovering Semantics and Topics Utilized by Threat Actors to Deliver Malicious Attachments and URLs}

\renewcommand{\shorttitle}{Uncovering Semantics and Topics Utilized by Threat Actors}

%%
%% The "author" command and its associated commands are used to define
%% the authors and their affiliations.
%% Of note is the shared affiliation of the first two authors, and the
%% "authornote" and "authornotemark" commands
%% used to denote shared contribution to the research.
\author{Andrey Yakymovych}
\email{andrey@inceptioncyber.ai}

\author{Abhishek Singh}
\email{abhishek@inceptioncyber.ai}

%%
%% The abstract is a short summary of the work to be presented in the
%% article.
\begin{abstract}
Recent threat reports highlight that email remains the top vector for delivering malware to endpoints. Despite these statistics, detecting malicious email attachments and URLs often neglects semantic cues—linguistic features and contextual clues. Our study employs BERTopic unsupervised topic modeling to identify common semantics and themes embedded in email to deliver malicious attachments and call-to-action URLs. We preprocess emails by extracting and sanitizing content and employ multilingual embedding models like BGE-M3 for dense representations, which clustering algorithms (HDBSCAN and OPTICS) use to group emails by semantic similarity. Phi-3-Mini-4K-Instruct facilitates semantic and hLDA aid in thematic analysis to understand threat actor patterns. Our research will evaluate and compare different clustering algorithms on topic quantity, coherence, and diversity metrics, concluding with insights into the semantics and topics commonly used by threat actors to deliver malicious attachments and URLs, a significant contribution to the field of threat detection.
\end{abstract}

%%
%% Keywords separated with commas.

%%
%% This command processes the author and affiliation and title
%% information and builds the first part of the formatted document.
\maketitle

\section{Introduction}
The HP Q1 2024 Threat Report \cite{hp2024threat} highlights that 53\% of malware is delivered via email, with 12\% of these threats bypassing detection technologies to reach endpoints. Similarly, the 2024 Verizon Data Breach Report \cite{verizon2024threat} indicates that around 35\% of ransomware infections originated from email. Despite this, semantics—linguistic features and contextual clues—remain underutilized in detecting malicious email attachments and URLs.

\begin{figure}[h]
  \centering
  \includegraphics[width=.8\linewidth]{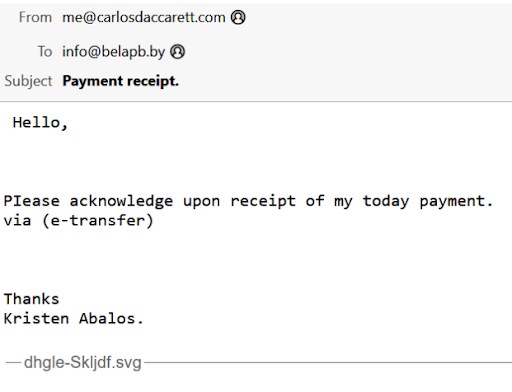}
  \caption{Email delivering malicious SVG.}
  \label{fig:svg_email}
  \Description{Email with subject 'Payment receipt' and the following text body: Hello, PIease acknowledge upon receipt of my today payment. via (e-transfer) Thanks Kristen Abalos.}
\end{figure}

Let's take an example to better understand how semantics and thematic embedded in emails can be used as a feature set to determine malicious attachment or URL. As demonstrated in Figure \ref{fig:svg_email} by a recent GUloader campaign that distributed malware through malicious SVG files delivered via email.

These SVG files, once downloaded, contained ZIP files with WSF scripts. The WSF scripts executed PowerShell commands to connect to a malicious domain and run shellcode injected into the MSBuild application. The attackers used semantics related to payment receipts, requesting acknowledgments to deliver the malicious SVG files. By leveraging learning around semantic analysis, the attachment could have classified the attachment as malicious without knowing the subsequent attack stage. This practical example underscores the value of understanding the semantic and thematic meaning embedded in the body of an email which then can be used as a feature set in real-world threat detection scenarios to classify attachment or URL as malicious.

In this research, we analyze past emails used to deliver malicious attachments and derive the semantics, themantics and topics extensively employed by the threat actors.

\begin{figure}[h]
  \centering
  \includegraphics[width=.8\linewidth]{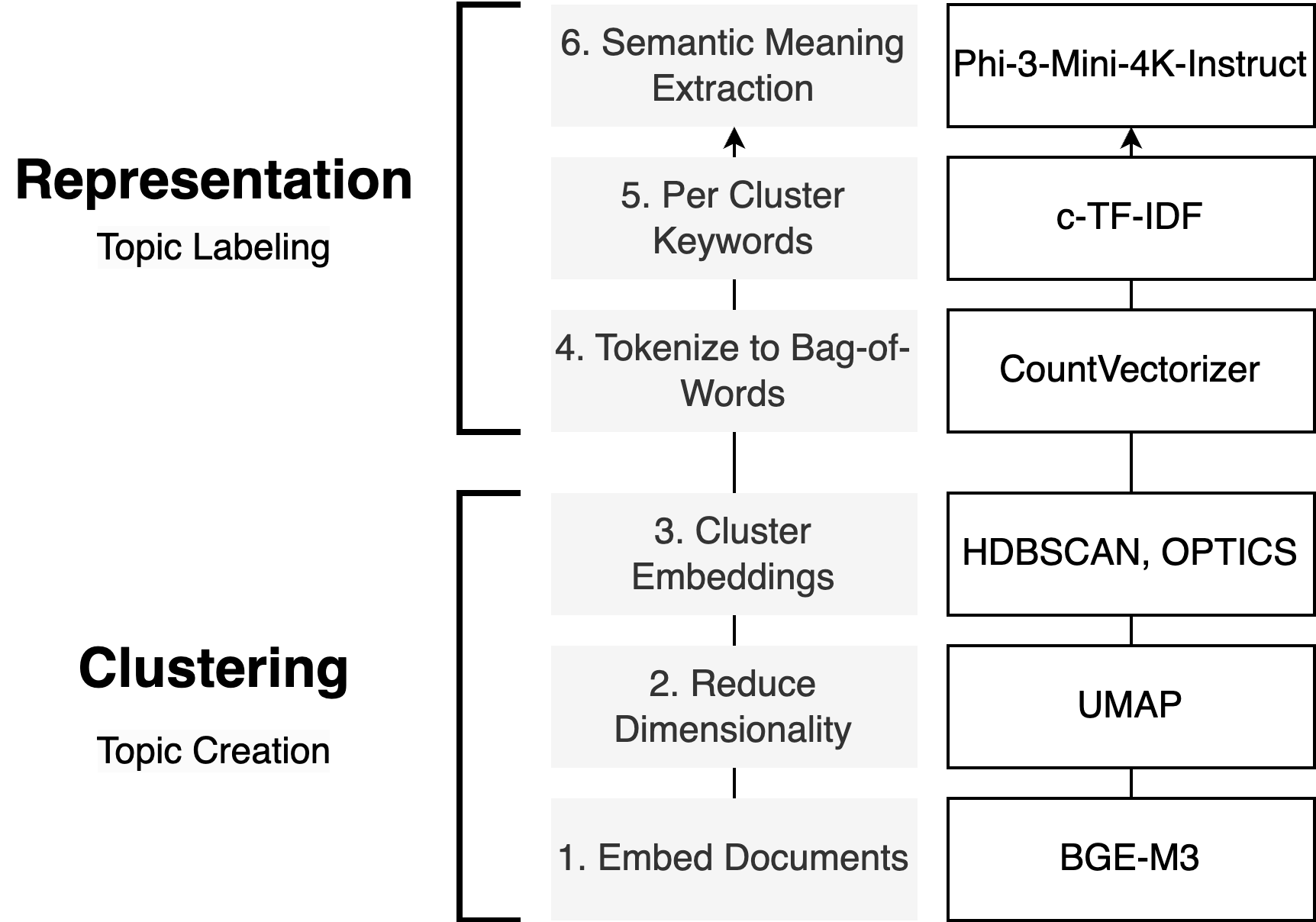}
  \caption{Diagram of the BERTopic pipeline showing the submodel configuration.}
  \label{fig:pipeline_config}
\end{figure}

\section{Experimental Setup}
BERTopic \cite{grootendorst2022bertopic} is a modular, unsupervised method of discovering topics in large collections of text and operates in two stages: topic creation and topic labeling.

To create topics, documents are first converted to an embedding representation using a pre-trained embedding model and dimensionality of the embeddings is reduced to enable effective clustering. A clustering algorithm then processes these reduced embeddings to group documents by semantic similarity. The selection and configuration of the clustering algorithm is the primary factor that determines the quantity and quality of topics formed from a set of embeddings.

To label topics, the aggregated text of each cluster is first tokenized to a bag-of-words representation. Each topic’s representative keywords are obtained through class-based Term Frequency Inverse Document Frequency (c-TF-IDF), which modifies TF-IDF to consider the cluster as a single document. Finally, this keyword representation is supplemented with the semantic meaning of the cluster’s documents extracted via Phi-3-Mini-4K-Instruct \cite{abdin2024phi3technicalreporthighly}.

\subsection{Datasets}
We collected 81360 malicious email samples that spanned multiple languages through VirusTotal’s dataset. These were then processed by our custom parser. This dataset was mixed with virus alert responses which were generated post-infection and were not representative of the intent of threat actors. As such, these were filtered out by checking for “virus”, “spam”, “alert” keywords in the subject.

Text entries were formed from each sample by extracting text from the email body and prepending the subject. Text cleaning was performed by removing large alphanumeric and symbol sequences as we found this would diminish clustering performance and pollute topic representations. Based on our analysis of malicious emails, we decided to omit text entries exceeding 7000 characters in length from this dataset as this was near the average length of the text body and mitigated the computational cost of extremely large samples. The remaining 67644 samples were used.

The document embedding and topic representation stages in BERTopic are separated which allows for flexible preprocessing \cite{grootendorst2022bertopic}. The full-length text entries were used at the embedding stage to retain as much information as possible for clustering. At the semantic meaning extraction stage, each text entry was truncated to 500 characters in order to fit within the Phi-3-Mini-4K-Instruct context window and environment memory constraints.

\subsection{BERTopic Configuration}
BGE-M3 \cite{chen2024bgem3} was selected as an embedding model as it exhibits state-of-the-art multilingual and cross-lingual performance well-suited to our dataset. Embeddings were precomputed on a GPU. This initial representation would yield poor clustering quality and runtime performance as it has been shown that in high dimensional space (1024-dimensional for BGE-M3), distance and the concept of proximity are ineffective for measuring similarity \cite{aggarwal2001surprising}. UMAP is used to reduce the dimension of embeddings before clustering to overcome the curse of dimensionality as it has been shown to preserve the global structure of high-dimensional data well \cite{mcinnes2018umap}.

\begin{figure}[h]
  \centering
  \begin{subfigure}{.45\linewidth}
    \includegraphics[width=\textwidth]{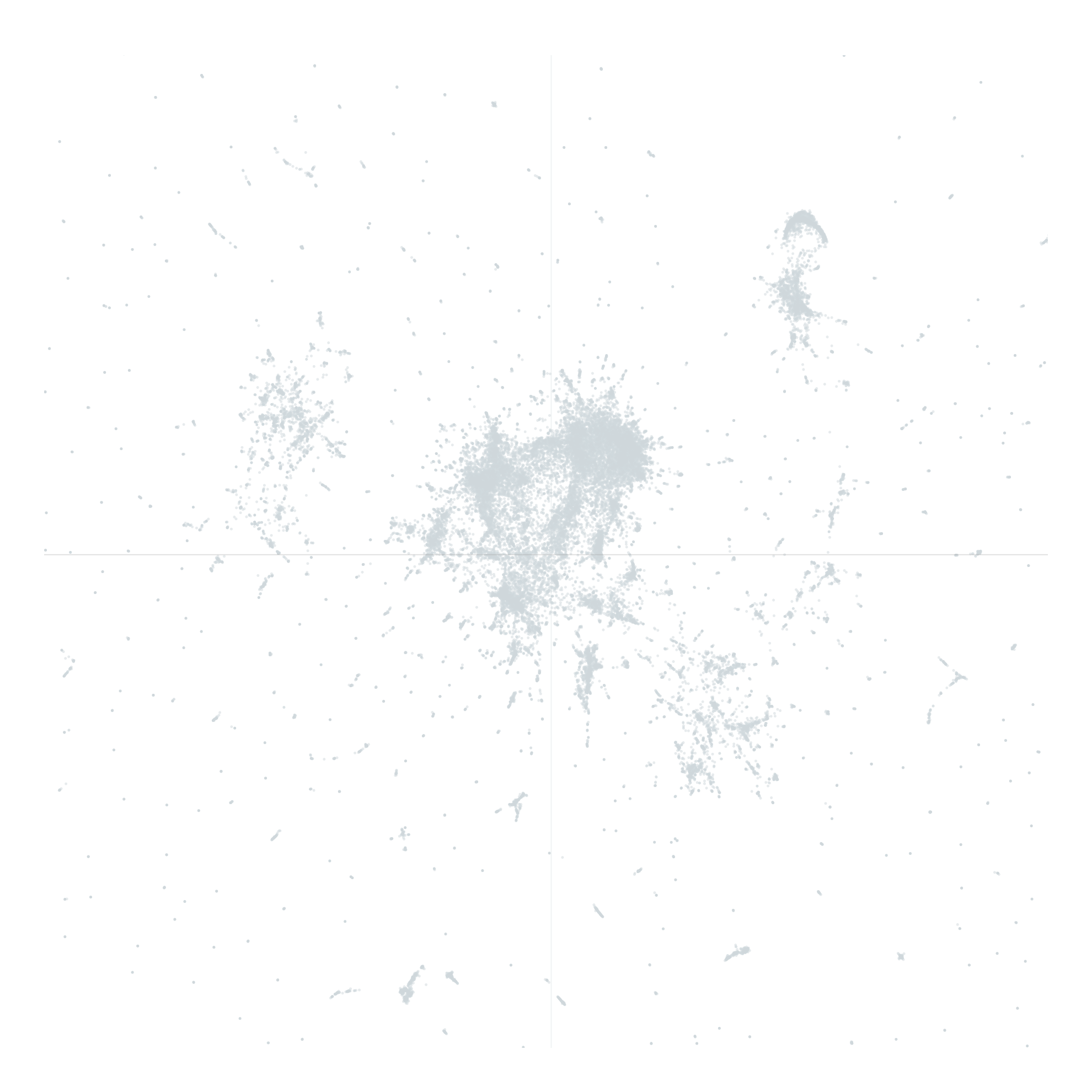}
    \caption{}
  \end{subfigure}
%%%%%%%%%%%%%%
  \begin{subfigure}{.45\linewidth}
    \includegraphics[width=\textwidth]{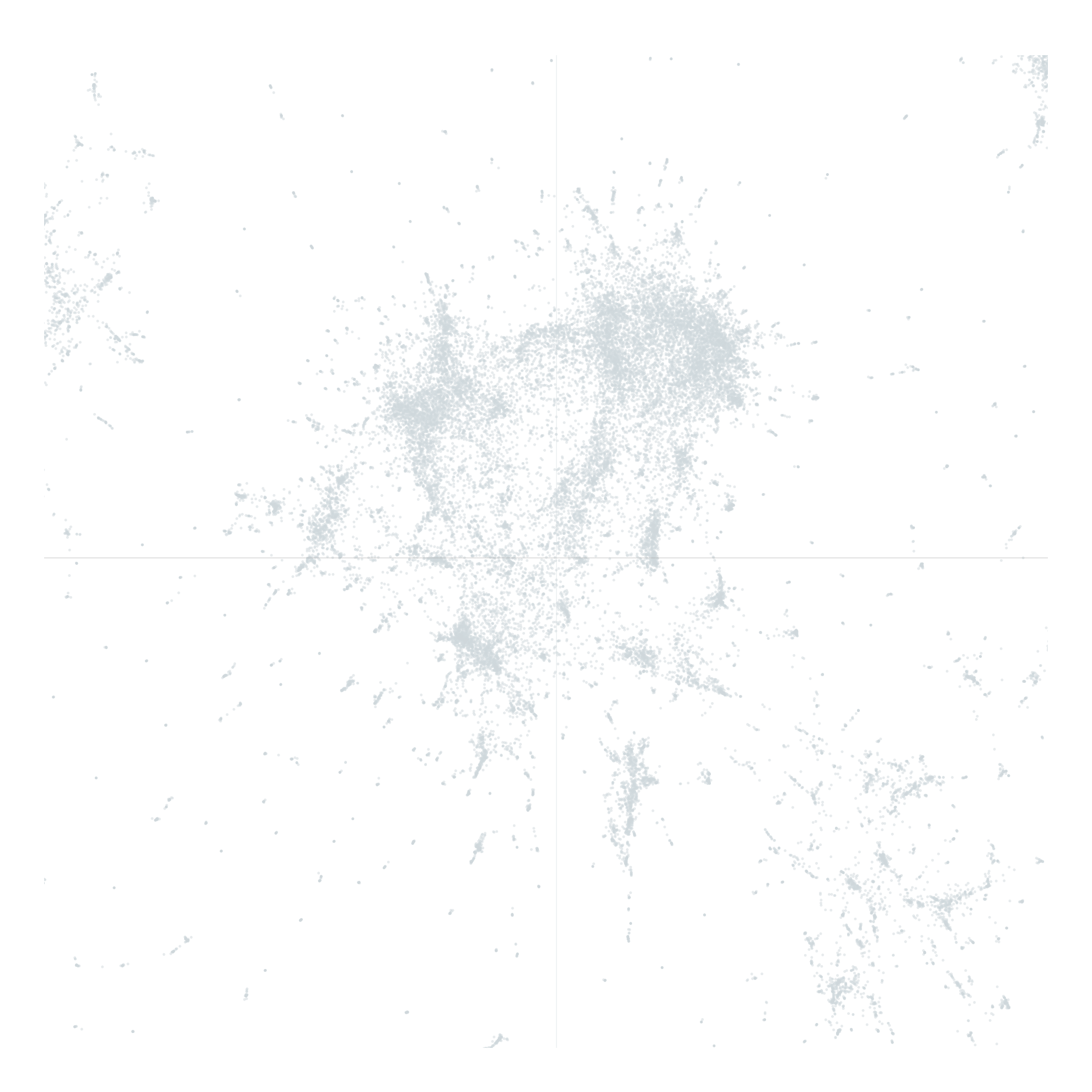}
    \caption{}
  \end{subfigure}
  \caption{Visualizations of BGE-M3 embeddings at multiple zoom levels. These show the variable density and complex shapes of potential clusters.}
  \label{fig:embedding_visual}
\end{figure}

The primary point of experimentation was in the choice of clustering algorithm and its hyperparameters as this would dictate the quantity and variety of topics formed. Based on our analysis of the embeddings (Figure \ref{fig:embedding_visual}), we determined that the hierarchical density-based algorithms, HDBSCAN and OPTICS, would be best suited for clustering. Density-based algorithms such as these have been shown to discover arbitrarily-shaped clusters better than centroid-based clustering whilst also detecting outliers. The hierarchical aspect of the chosen algorithms enables them to discover clusters of varying density \cite{mcinnes2017hdbscan} by not requiring the epsilon distance parameter of the related DBSCAN algorithm. HDBSCAN and OPTICS were trialed with varying minimum cluster sizes (50, 100, 150) and cluster methods (HDBSCAN: excess of mass, leaf; OPTICS: xi).

Phi-3-Mini-4K-Instruct was used to extract semantic meaning as it is a lightweight state-of-the-art model \cite{abdin2024phi3technicalreporthighly}.

\subsection{Evaluation}

Clustering methods were evaluated on a combination of three metrics: topic quantity, topic coherence, and topic diversity.

Topic coherence was evaluated using normalized pointwise mutual information (NPMI \cite{bouma2009normalized}) which gauges the interpretability of a topic by how closely associated the most common keywords in each topic are. Coherence close to -1 indicates poor association and 1 indicates strong association.

Topic diversity, as defined by Dieng et al. \cite{dieng2020topic}, is the percentage of unique words for all topics. Diversity close to 0 indicates redundant topics and 1 indicates more varied topics. Dieng et al. also defined topic quality as the product of topic coherence and topic diversity. Higher values for this metric are better and indicate that the topics produced by a clustering method are both interpretable and varied.

Topic quantity alone is a naive measurement of the granularity of a clustering method, as it fails to capture the interpretability and variety of the topics produced. As such, we compute the product of topic quantity and topic quality as an overall metric, higher values of which measure the granularity, interpretability and variety of the topics produced by a clustering method.

Evaluation was conducted using the Gensim coherence model for NPMI coherence and OCTIS (Optimizing and Comparing Topic models Is Simple) \cite{terragni2021octis} for topic diversity.

\begin{figure}[h]
  \centering
  \begin{subfigure}{.45\linewidth}
    \includegraphics[width=\textwidth]{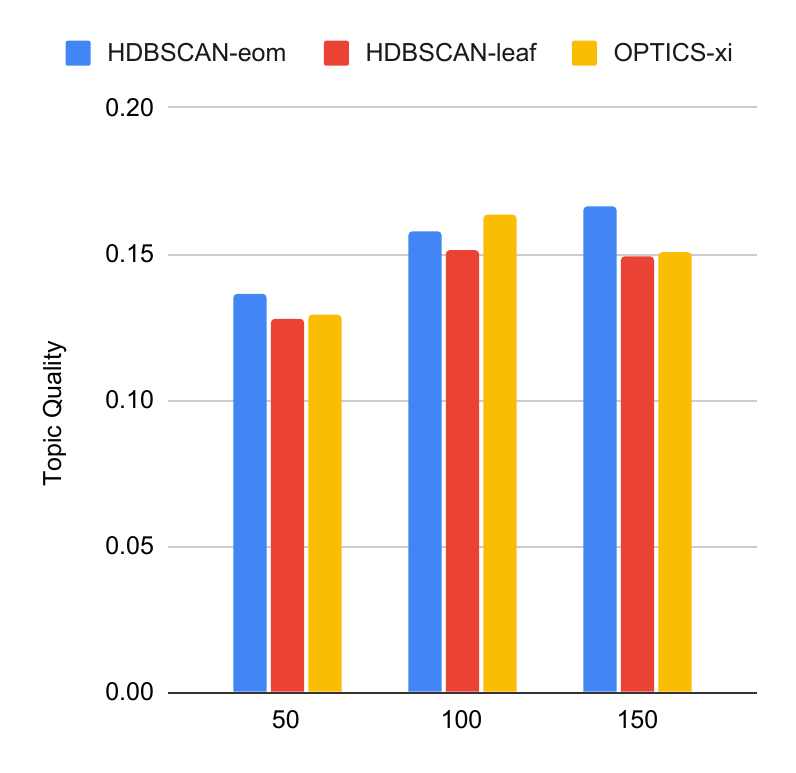}
    \caption{Topic Quality}
  \end{subfigure}
%%%%%%%%%%%%%%
  \begin{subfigure}{.45\linewidth}
    \includegraphics[width=\textwidth]{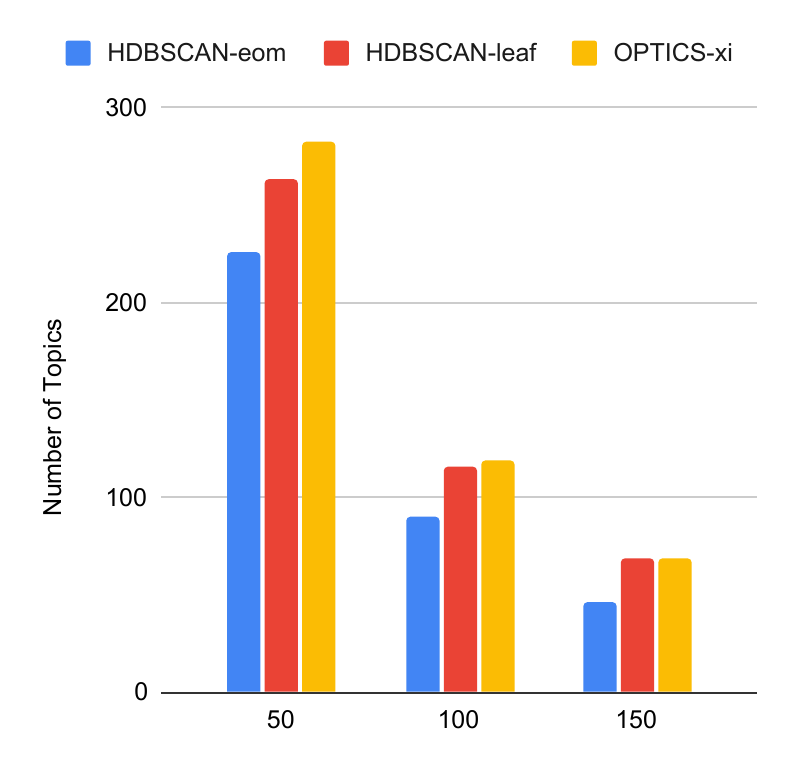}
    \caption{Number of Topics}
  \end{subfigure}
%%%%%%%%%%%%%%
  \begin{subfigure}{.45\linewidth}
    \includegraphics[width=\textwidth]{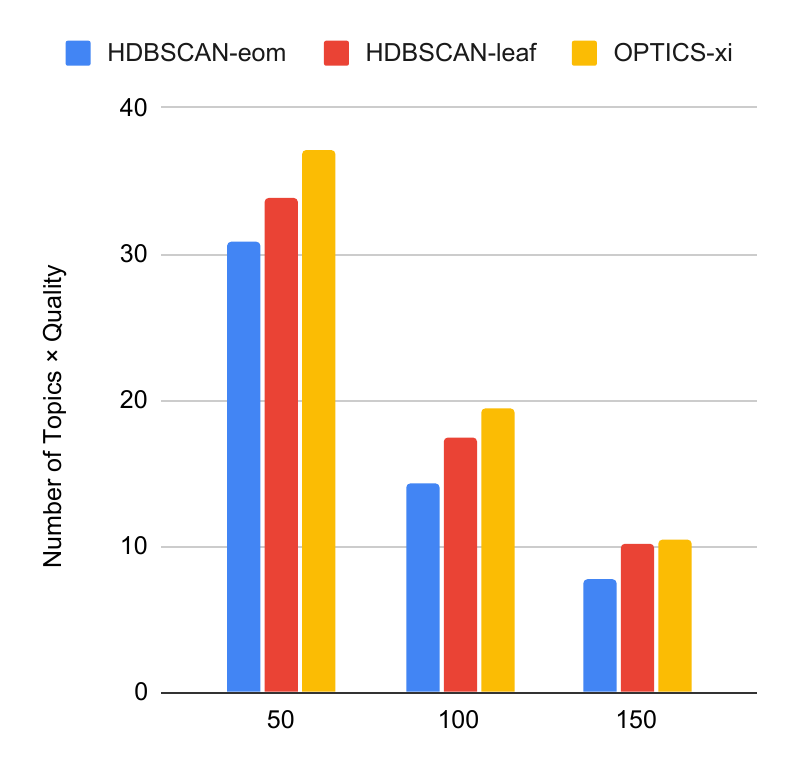}
    \caption{Number of Topics \texttimes{} Quality}
  \end{subfigure}
  \caption{Evaluation results of clustering configurations grouped by minimum cluster size. Scores were averaged over 3 runs for each clustering configuration. One of the OPTICS-xi (minimum cluster size: 50) runs produced an error in the coherence metric calculation, but this didn’t affect the conclusion.}
  \label{fig:evaluation_charts}
\end{figure}

\section{BERTopic Clustering Results}
Our BERTopic clustering results can be found in Figure \ref{fig:evaluation_charts}. Scores were averaged over 3 runs for each clustering configuration. Run-to-run variance and reproducibility were controlled by setting a random seed in the UMAP dimension reduction.

\subsection{Performance}
From Figure \ref{fig:evaluation_charts}, we can observe that OPTICS-xi clustering generally produces 25.2\% more topics than HDBSCAN-eom whilst maintaining competitive quality of 94.9\% of HDBSCAN-eom’s quality. HDBSCAN-leaf has similar performance differences when compared to HDBSCAN-eom but only produces 16.8\% more topics with a lower relative quality of 94.0\%.

Increasing the minimum cluster size hyperparameter generally results in improved topic quality, but reduced topic quantity. Increasing this hyperparameter from 50 to 100 for instance, increases the topic quality metric by 20.1\% but reduces the number of topics by 58.0\%.

\section{Analysis}
Since each cluster is composed of malicious emails with similar semantic meaning, the formation of distinct clusters indicates that topics and semantics are repeated by threat actors. Since our dataset and embedding model are multilingual, this repetition of semantic patterns was also consistently identified in non-English languages.

For detailed analysis, we selected a run of OPTICS-xi with a minimum cluster size of 50 as it produced a high topic quantity whilst retaining competitive topic quality. This run formed 295 clusters whose semantics were representative of 14 topic categories commonly used by threat actors across multiple languages.

\begin{figure}[H]
  \centering
  \includegraphics[width=\linewidth]{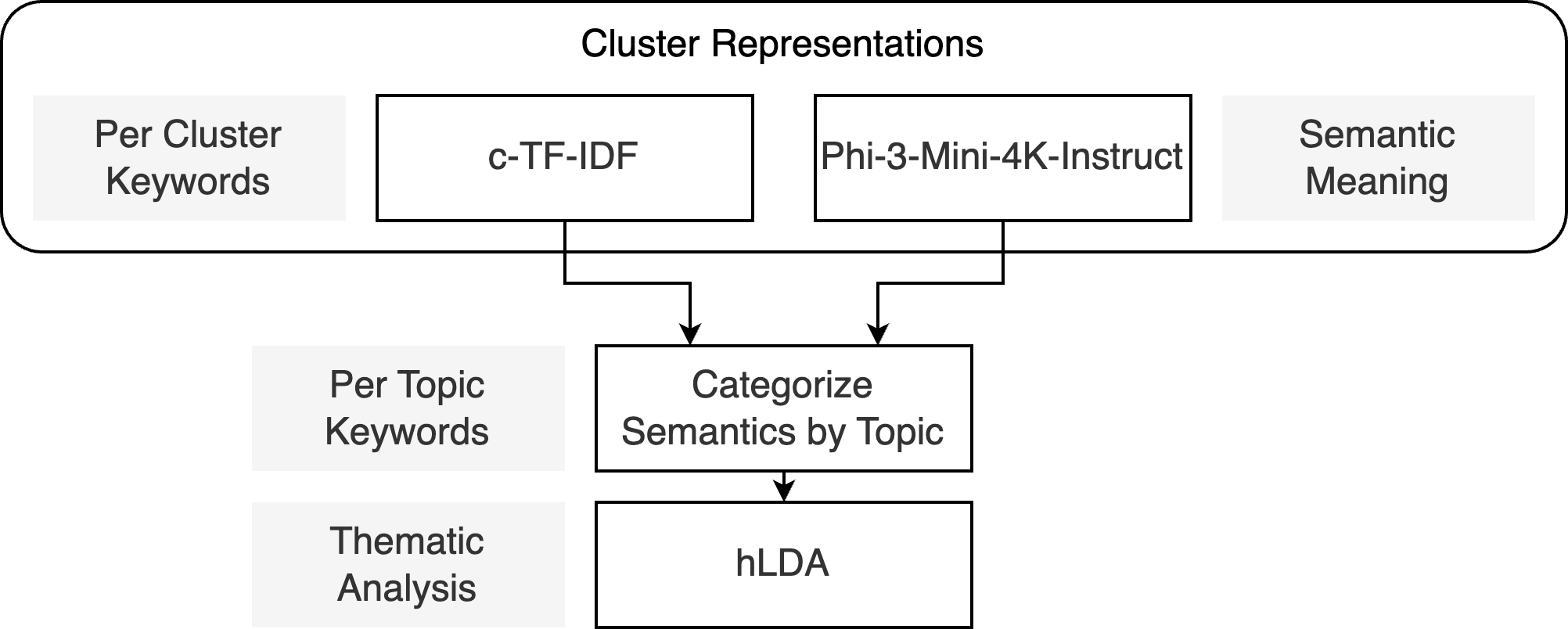}
  \caption{Workflow for semantic and thematic analysis.}
  \label{fig:semantic_analysis_workflow}
\end{figure}

The semantic meaning extracted by Phi-3-Mini-4K-Instruct facilitates identifying topic categories by grouping clusters with related semantics. Each cluster’s c-TF-IDF keyword representation is used to form a keyword dictionary organized by topic category. This dictionary of keywords can be used for thematic analysis, as each cluster can contain a combination of topics and hierarchical structure of topics can be formed.

\begin{table}[h]
  \caption{Keywords and semantics for a cluster of the financial topic category. Samples of this cluster have semantics that prompt the recipient to open a malicious archive under the pretense that it contains a bank account report.}
  \label{tab:financial_thematic_analysis}
  \begin{tabular}{p{0.22\linewidth}p{0.22\linewidth}p{0.22\linewidth}p{0.22\linewidth}}
    \hline
    Name & c-TF-IDF Keyword Representation & Phi-3-Mini-4K-Instruct Semantic Meaning & Topic Hierarchy / Thematic Analysis\\
    \hline
    {financial responding disapproval statementhi} & ['financial', 'responding', 'disapproval', 'statementhi', 'short', 'king', 'very', 'topic', 'reporthello', 'monthly'] & Monthly Financial Response Evaluation Processing & {'financial': ['informational']}\\
    \hline
  \end{tabular}
\end{table}

\begin{figure}[H]
  \centering
  \includegraphics[width=.5\linewidth]{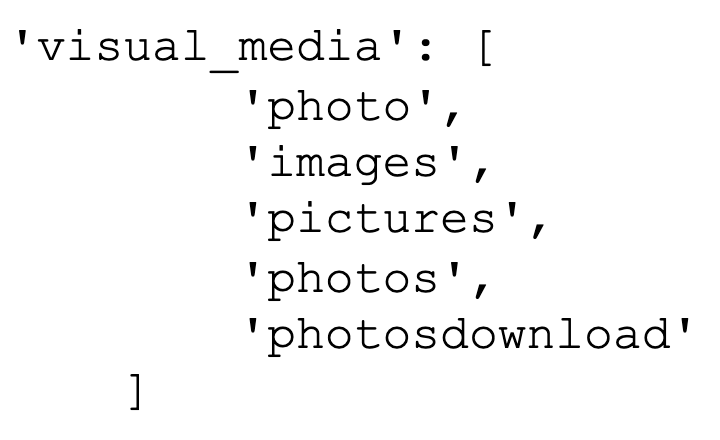}
  \caption{Keywords identified for the visual media topic category by aggregating keywords from clusters with semantics related to downloading or viewing images.}
  \label{fig:visual_media_category}
  \Description{'photo', 'images', 'pictures', 'photos', 'photosdownload'}
\end{figure}

\begin{figure*}[h]
  \centering
  \includegraphics[width=\linewidth]{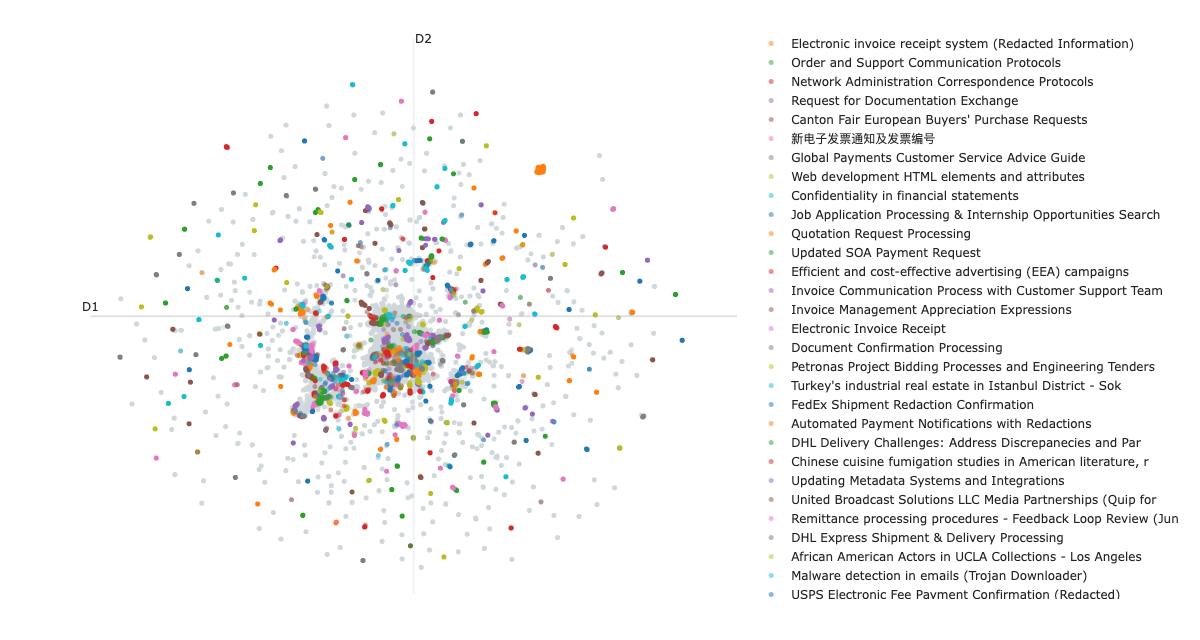}
  \caption{Malicious email clustering visualization obtained with an OPTICS-xi (minimum cluster size: 50) configuration of BERTopic.}
  \label{fig:clustering_visual}
  \Description{}
\end{figure*}

\subsection{Thematic Analysis}
Hierarchical topic modeling generates a hierarchical structure of topics, mapping primary topics to subtopics, thereby effectively organizing thematic content within the text into a structured representation. In hierarchical topic modeling, BERTopic excels in capturing semantic similarity, while hLDA is more effective at identifying a larger number of distinct topics and subtopics within a corpus, leading to a better understanding of the themes used by threat actors. Other algorithms, such as the Hierarchical Dirichlet Process (HDP), were also experimented with for thematic analysis. However, since HDP relies on random samples and a probabilistic approach, the output did not produce stable or interpretable topic hierarchies for the same dataset when executed multiple times. These factors made hLDA a suitable algorithm for hierarchical topic modeling to capture the themes embedded in the emails delivered by threat actors.

Hierarchical Latent Dirichlet Allocation (hLDA \cite{blei2004heirarchical}) was employed to extract hierarchical topics from the given text, utilizing 15 primary topics and iterating through 50 passes. Initially, the text undergoes preprocessing, including converting it to lowercase, tokenizing, and removing stop words and punctuation. A dictionary of unique tokens is created from the preprocessed text, and the text is represented as a bag-of-words. If an LDA model is not already initialized, it is set up at this stage; otherwise, the existing model is updated with the new corpus. Topics are extracted from the model using its topic printing method.

These extracted topics are then matched to predefined categories based on a keyword dictionary. This process involves identifying primary topics and their corresponding sub topics by analyzing the keywords associated with each topic. This context helps in capturing the relationships and dependencies between different aspects of the text, leading to more meaningful thematic analysis and topic representations. The use of 15 topics and iterating through the 50 passes ensures a detailed and comprehensive analysis for capturing a more accurate and nuanced topic hierarchy.

\section{Conclusions}
Our research delves into the details of an artificial intelligence-based approach to semantic and thematic analysis, uncovering deeper meanings embedded within email bodies, which threat actors often exploit to deliver malicious attachments such as ransomware, password stealers, and call-to-action URLs. Our findings reveal that both semantic and thematic elements are frequently reused by threat actors, establishing them as generic features that can be utilized in heuristic or artificial intelligence-based threat detection algorithms.

Through the analysis of historic emails from threat actors, our research demonstrates that multilingual embedding models, BGE-M3, excel in creating dense representations. Density-based clustering algorithm OPTICS effectively generates a large number of interpretable and varied clusters, making it suitable for grouping emails based on semantic similarity. Additionally, the Phi-3-Mini-4K-Instruct model significantly enhances semantic understanding and aids in the generation of semantics used by threat actors. In the realm of hierarchical topic modeling, hLDA proves highly effective for thematic analysis, aiding in comprehending the tactics of threat actors.

%%
%% The next two lines define the bibliography style to be used, and
%% the bibliography file.
\bibliographystyle{ACM-Reference-Format}
\bibliography{references.bib}

\clearpage
\appendix
\section{Additional Semantic and Thematic Analysis}
In the section, we present additional semantic and thematic analysis for 3 clusters of the OPTICS-xi (minimum cluster size: 50) run.

\begin{table}[H]
  \caption{More detailed analysis of 3 clusters chosen from 3 topics. The financial cluster contains emails whose semantics ask the recipient to look at a bank account report that is actually a malicious archive. Samples of the digital communication cluster semantics feature a call-to-action for the recipient to open a malicious .rar archive under the pretense of an unchecked voicemail. Samples of the visual media cluster prompt the recipient to open malicious attachments that claim to contain photographs.}
  \label{tab:full_thematic_analysis}
  \begin{tabular}{p{0.12\linewidth}p{0.18\linewidth}p{0.22\linewidth}p{0.18\linewidth}p{0.16\linewidth}}
    \hline
    Topic Category & Name & c-TF-IDF Keyword Representation & Phi-3-Mini-4K-Instruct Semantic Meaning & Topic Hierarchy / Thematic Analysis\\
    \hline
    Financial & {financial responding disapproval statementhi} & ['financial', 'responding', 'disapproval', 'statementhi', 'short', 'king', 'very', 'topic', 'reporthello', 'monthly'] & Monthly Financial Response Evaluation Processing & {'financial': ['informational']}\\
    \hline
    Digital Communication & {mailbox just chance voicemail} & ['mailbox', 'just', 'chance', 'voicemail', 'wanted', 'might', 'voice', 'long', 'were', 'fri'] & Voicemail reminder notification (Fri) & {'digital communication': [], 'call to action': ['digital communication']}\\
    \hline
    Visual Media & {photomy photo pfoto recording} & ['photomy', 'photo', 'pfoto', 'recording', 'new', 'missed', 'my', 'voicemail', 'containing', 'invoices'] & Personal photography updates with details & {'invoice': ['visual media', 'call to action', 'informational', 'digital communication']}\\
    \hline
  \end{tabular}
\end{table}

\section{GUloader Campaign Samples}

Samples of GUloader campaign can be retrieved from VirusTotal using the provided SHA-256 file hash.

Malicious SVG: \hash{b20ea4faca043274bfbb1f52895c02a15cd0c81a333c40de32ed7ddd2b9b60c0}

Malicious Email: \hash{66b04a8aaa06695fd718a7d1baa19386922b58e797634d5ac4ff96e79584f5c1}

\end{document}